\documentclass[conference]{IEEEtran}
\pdfoutput=1
\usepackage{times}
\usepackage{epsfig}
\usepackage{graphicx}
\usepackage{amsmath}
\usepackage{amssymb}
\usepackage[margin=0.85in,bottom=0.9in,top=1in]{geometry}
\usepackage{subfigure}
\usepackage[none]{hyphenat}
\usepackage{epstopdf}
\usepackage[numbers,sort&compress]{natbib}
\usepackage{fancyhdr}
\begin{document}
\title{Identification of Internal Faults in Indirect Symmetrical Phase Shift Transformers Using Ensemble Learning}
\author{Pallav Kumar Bera\\
Syracuse University\\
New York, USA\\
{\tt\small pkbera@syr.edu}
\and
Rajesh Kumar\\
Syracuse University\\
New York, USA\\
{\tt\small rkuma102@syr.edu}
\and
Can Isik\\
Syracuse University\\
New York, USA\\
{\tt\small cisik@syr.edu}}
\IEEEoverridecommandlockouts
\IEEEpubid{\makebox[\columnwidth]{978-1-5386-7568-7/18/\$31.00 \copyright 2018 IEEE \hfill} \hspace{\columnsep}\makebox[\columnwidth]{ }}
\maketitle
\IEEEpubidadjcol
\begin{abstract}
This paper proposes methods to identify $40$ different types of internal faults in an Indirect Symmetrical Phase Shift Transformer (ISPST). The ISPST was modeled using Power System Computer Aided Design (PSCAD)/ Electromagnetic Transients including DC (EMTDC). The internal faults were simulated by varying the transformer tapping, backward and forward phase shifts, loading, and percentage of winding faulted. Data for $960$ cases of each type of fault was recorded. A series of features were extracted for a, b, and c phases from time, frequency, time-frequency, and information theory domains. The importance of the extracted features was evaluated through univariate tests which helped to reduce the number of features. The selected features were then used for training five state-of-the-art machine learning classifiers. Extremely Random Trees and Random Forest, the ensemble-based learners, achieved the accuracy of $98.76\%$ and $97.54\%$ respectively outperforming Multilayer Perceptron ($96.13\%$), Logistic Regression ($93.54\%$), and Support Vector Machines ($92.60\%$). 

\end{abstract}


\section{Introduction}
Increased real power requirements eventually result in increased reactive power demands and voltage variation along the transmission lines. Although reactive shunt compensation and on load tap changers meet reactive power requirement, system voltage may collapse in case of on load tap changers. Series reactive compensation handles real power flow issues properly but fails in cases related to the transmission angle. For example, incompatibility of the prevailing transmission angle with the transmission requirements and control of real and reactive power flow in meshed networks. Phase Shift Transformers (PSTs), introduced in the 1930’s, are used to mitigate the problems involving power transmission angle. In addition, they can be used to control reactive and real power with quadrature and in-phase voltage regulation. Modern phase angle regulators can also control dynamic events like transient instability, damp out power oscillations (when $d\delta/dt>0$ phase shift is made negative and when $d\delta/dt<0$ it is made positive), post-disturbance overloads and subsequent voltage dips and power oscillations \cite{Facts}. A PST controls the real power flow by controlling the $(\delta)$. The modified real power flow (P) in a transmission line with a PST is specified in equation (1).
\begin{equation}
    P=\frac {V_s \times V_l}{X_l+ X_{pst}} \times sin(\delta \pm \alpha)
\end{equation}
where, $\alpha$ is change in phase angle, $V_s$ and $V_l$ are the  sending and receiving end voltages respectively, and $X_l$ and $X_{pst}$ are the impedance of transmission line and PST respectively \cite{Harlow}.

\begin{figure}[htp]
\centerline{\includegraphics[width=2.8 in, height=2.4 in]{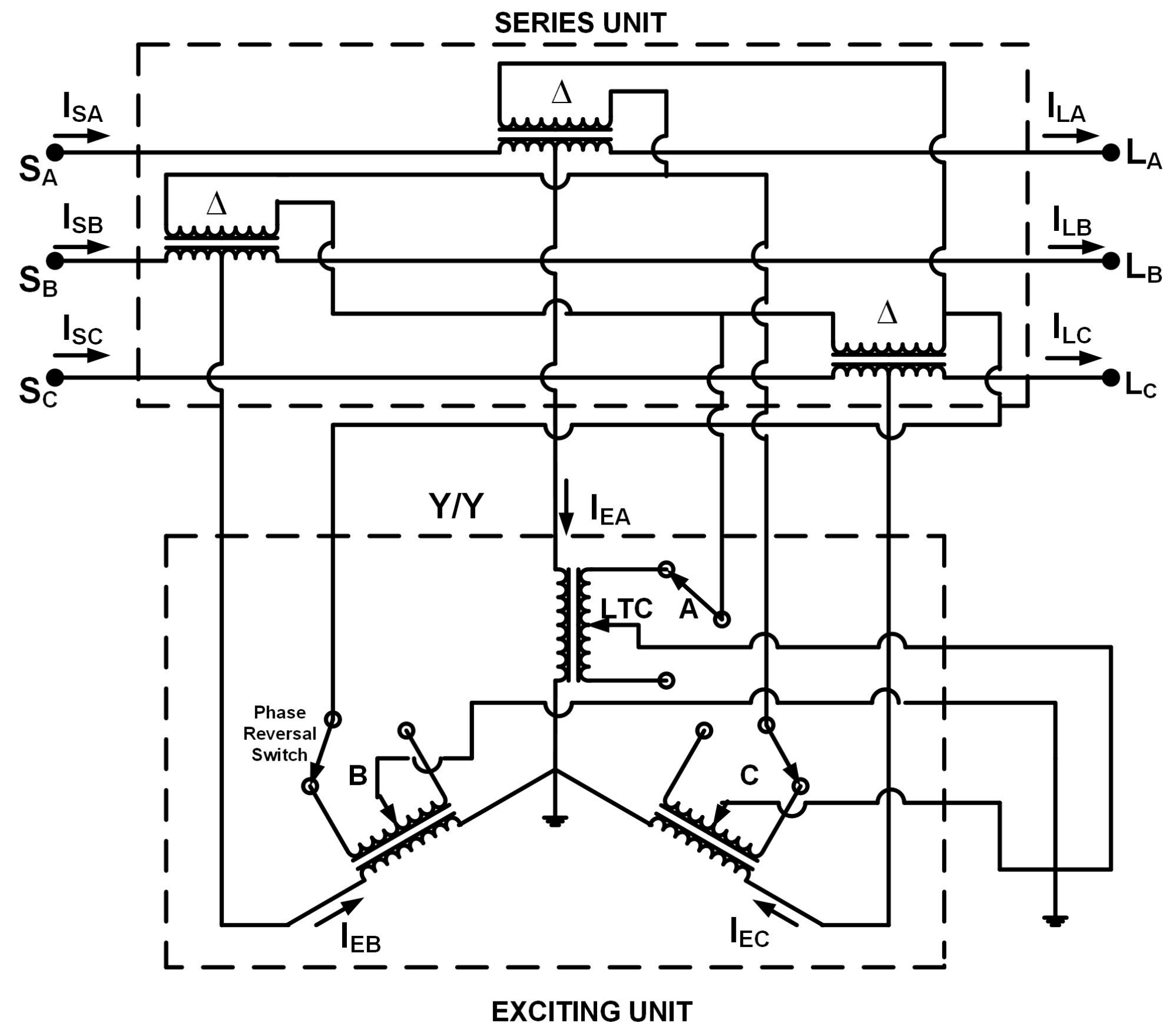}}
\caption{An ISPST with the secondary of series transformer unit connected in delta and the exciting transformer unit having LTCs and phase reversal switches (on secondary) connected in star-star with both neutrals grounded. \cite{ibrahim}}
\label{ispst}
\end{figure}

Depending on the number of transformer units and the magnitude of the output voltage, PSTs are categorized into four groups. Direct PSTs have a single 3-phase transformer unit. Indirect PSTs have two 3-phase transformer units, an exciting transformer with a tap changer to adjust the amplitude of the quadrature voltage and a series transformer which injects the quadrature voltage in the required phase. Asymmetrical PSTs have an output voltage which has a different phase angle and amplitude than the input voltage, whereas, symmetrical PSTs have the same output and input voltages but with a different phase angle. Indirect Symmetrical PSTs (ISPST) are widely used since they offer greater operational security at high voltages as the load tap changer (LTC) is not exposed to system disturbances. Therefore, this paper focuses on ISPST. Figure \ref{ispst} shows an ISPST with two 3-phase transformer units.

Information about the type of internal fault in ISPST is required for the identification of the faulty unit, evaluation of the amount of repair work needed, and fault analysis. Several researchers have proposed various intelligent techniques for protection and classification of faults in Power Transformers. For example, Tripathy et al. used Probabilistic Neural Network (PNN) to detect different conditions such as magnetizing inrush, over-excitation, internal and external faults in Power Transformer operation \cite{tripathy}. Similarly, Balaga et al. implemented Transformer protection using genetic algorithm based parallel hidden layered Artificial Neural Network (ANN) \cite{balaga}. Mittal et al. \cite{mittal}, Bigdeli et al. \cite{bigdeli} and Koley et al. \cite{koley} proposed Support Vector Machine (SVM) and ANN based fault classification. Patel et al. \cite{patel} suggested that Relevance Vector Machine (RVM) performs better than PNN and SVM. Although enough literature exists which support the use of machine learning techniques in protection and classification of faults in Power Transformers very few have been realized for protection and classification of faults in PST. Even though fewer in number, PSTs play a vital role in a power system and are very expensive. With two 3-phase transformers units, ISPST offers a challenging internal fault identification problem. Considering the primary and secondary sides of these two transformers, identification of internal faults becomes more complex and extensive than in the case of Power Transformers. Wavelet transform was used for protection of ISPST by Bhasker et al \cite{paper2}. Further, they used two Multi-Layer Feed Forward Neural Networks (MLFFNN) to classify different internal faults \cite{pallav}. The first network was used to identify the type and the second to identify the side of the faulty transformer unit. However, these studies were limited to 20 different faults in a-phase to ground (a-g) for line to ground (l-g), a-phase to b-phase to ground (ab-g) for line to line to ground (ll-g), and b-phase to c-phase (bc) for line to line (ll) on the primary and the secondary sides of the two transformer units. On the other hand, this paper studies 40 different internal faults that cover all three phases (a, b, and c). The main contributions of this paper are as follows:
\begin{itemize}
    \item An ISPST was modeled in PSCAD. More specifically, two-winding and three-winding transformer components were designed which were absent in the PSCAD/EMTDC 4.2.0 library and were validated using RSCAD/RTDS.
    
    \item  Considering all three phases, a total of 40 different types of faults were simulated. A dataset of 38,400 cases, 960 for each type, was created. These many faults have not been studied together before.
    
    \item A series of features were extracted from time, frequency, time-frequency, and information theory domains. Further, univariate tests were conducted to identify and remove irrelevant features. The selected features were then supplied to five distinct classification algorithms for identification of the faults. 
    
    \item The performance of the algorithms were evaluated using accuracy, precision, recall, and f1-score. The dataset and Python scripts are publicly available.\footnote{https://github.com/rajeshjnu2006/ISPSTInternalFaults}
    
\end{itemize}

The rest of the paper is organized as follows: section II illustrates the modeling of the ISPST in PSCAD/EMTDC and data collection process for different internal faults. Section III talks about the theory behind the feature analysis. Section IV describes the classification framework used. Section V shows the simulation and classification results. Finally, section VI concludes the paper.

\section{Modeling and Fault Data Collection}
\begin{figure*}[htp]
\centering
\begin{tabular}{ccc}
\subfigure[ISPST model]{\epsfig{file=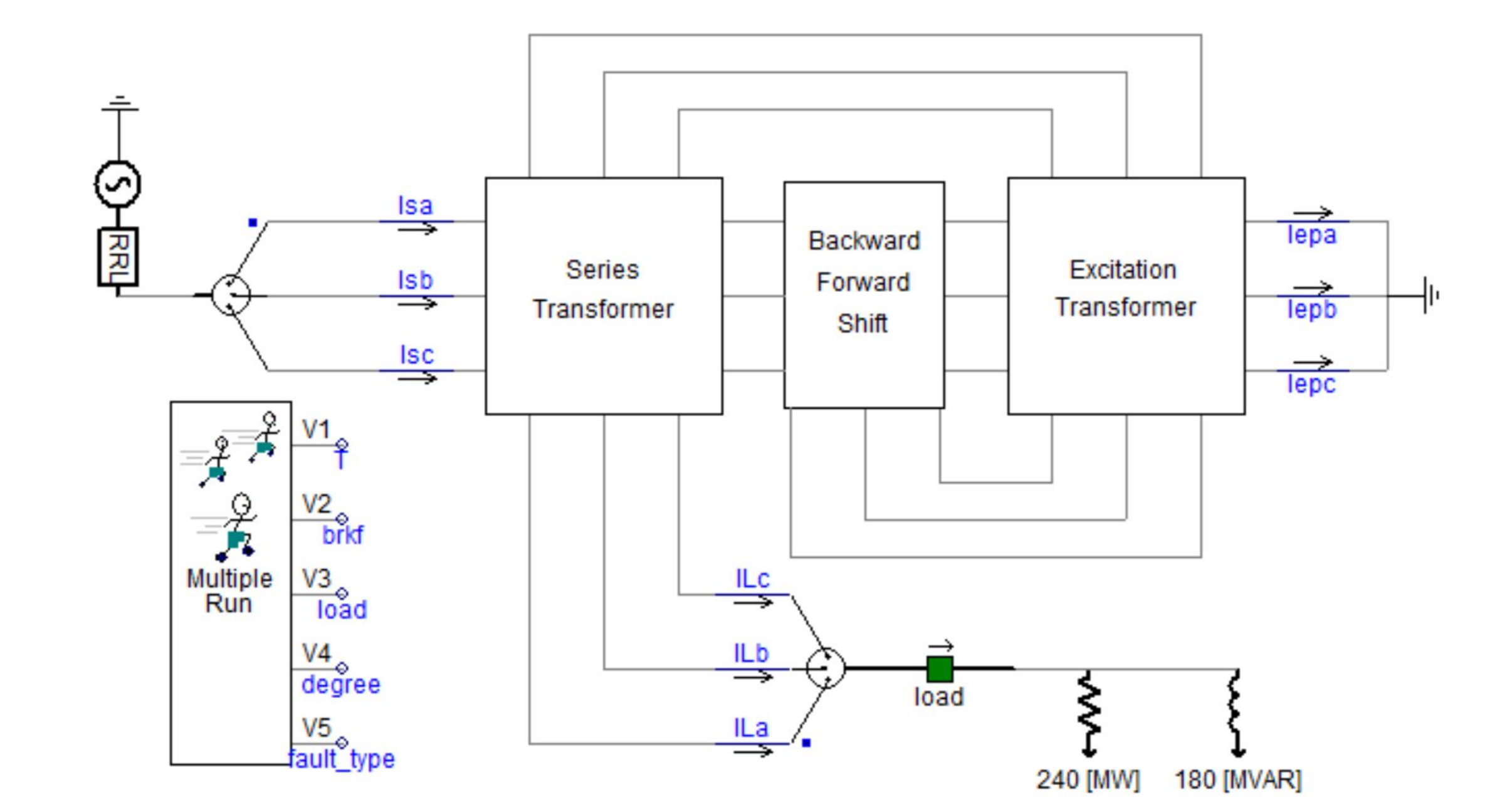, width=3 in, height= 1.8 in} 

\label{draw1fig1}}&
\subfigure[Series transformer unit]{\epsfig{file=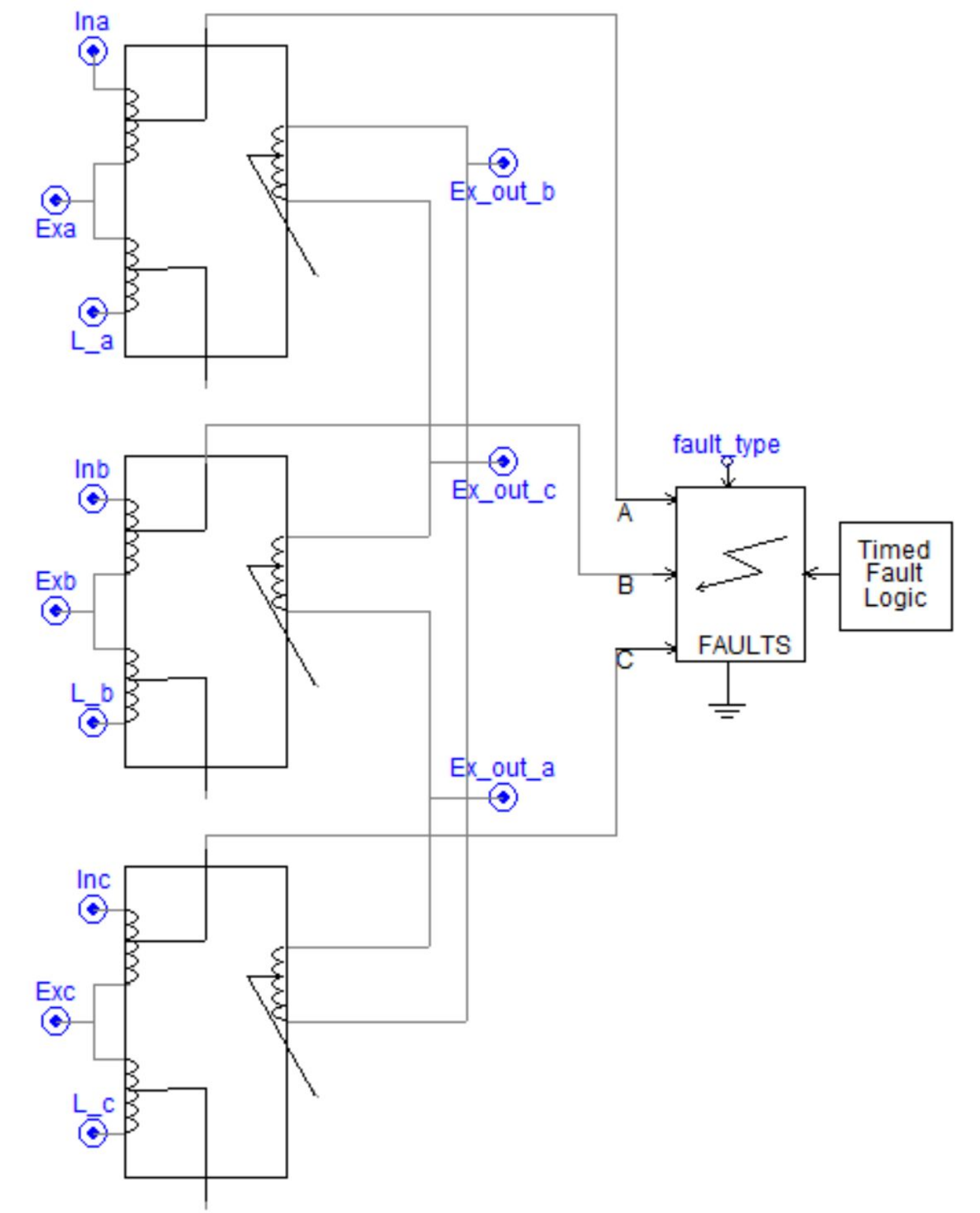, width=1.7 in, height=1.7 in}

\label{draw1fig2}}&
\subfigure[Exciting transformer unit]{\epsfig{file=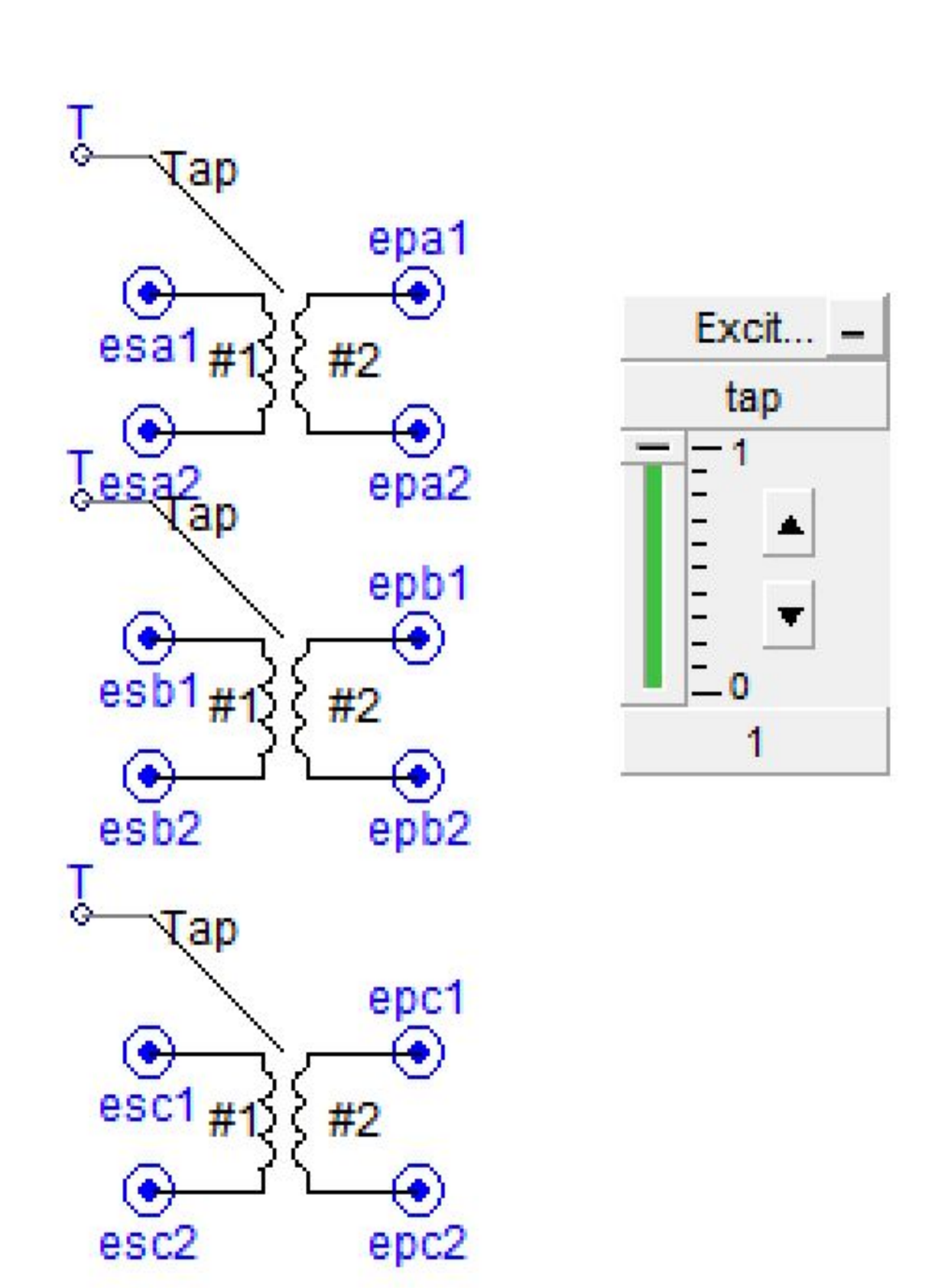, width=1.7 in, height=1.7 in}

\label{draw1fig3}}
\end{tabular}
\caption{A model of ISPST developed in PSCAD/EMTDC. Figure \ref{draw1fig1} represents the complete ISPST model with series, exciting, backward and forward phase shift circuit, ac source, inductive load, and multi-run components. Figure \ref{draw1fig2} is representing the series transformer consisting of 3-single phase three winding transformers from which internal faults on the primary side of the series transformer were generated. Figure \ref{draw1fig3} illustrates the exciting transformer consisting of 3-single phase two-winding transformers with the option to alter the tapping.}
\label{draw1fig}
\end{figure*}

PSCAD/EMTDC version 4.2.0 was used for the modeling and simulation of the ISPST. The ISPST was modeled using existing components in the PSCAD/EMTDC library. Additional required components, two and three winding transformers were designed using FORTRAN. The new components were validated by comparing them with the two and three winding transformer components of RSCAD/RTDS library. These components have provisions to alter leakage reactance between winding, saturation characteristics, and percentage of winding faulted. Figure \ref{draw1fig1} shows the developed model in PSCAD. The ISPST with a rated power of 300 MVA, rated voltage of 138 kV/138 kV, and rated current of 1.255 kA/1.255 kA was used to simulate the differential fault currents \cite{ibrahim}. The system frequency was kept at 60 Hz. The secondary side of the exciting transformer had an LTC which allowed phase angle shifts of $30$ degree forward to $30$ degree backward. The 3-phase series transformer was rated at 156.545 MVA ($3\times 41.579\times 1.255 $ MVA), 41.579 kV/61.783 kV. The 3-phase exciting transformer was rated at 149.986 MVA ($3\times 76.959\times 0.65 $ MVA), 76.959 kV/35.69 kV. Thus, every single phase exciting transformer was rated at 50 MVA and every series transformer was rated at 52.18 MVA. The 3-phase series transformer and  3-phase exciting transformers are shown in Figure \ref{draw1fig2} and Figure \ref{draw1fig3}.

L-g, ll-g, ll and 3- phase to ground faults were simulated using the multi-run component available in PSCAD/EMTDC library. The simulation run-time, fault inception time, and fault duration time were kept at 1.2 secs 0.5 secs, and 0.5 secs (3 cycles) respectively. Faults were simulated on the primary and secondary sides of the series and the exciting transformer units of the ISPST. The differential currents from the 3-phases (a, b, and c) were plotted at the rate of 100 micro-secs. Hence, an output file of 1201 rows and 4 columns was generated in each case. Column one represents time and columns 2, 3 and 4 represent the differential currents of the 3-phases. Three cycles of post-fault differential currents (endmost 700 samples out of 1201 samples) were captured from primary and secondary of the series and exciting transformer units. Four different parameters of the ISPST were varied to get data for training and testing. The inception angle was varied from $0$ degree to $345$ degree in steps of $15$ degree, the percentage of winding faulted was varied from 30\% to 70\% in steps of 10\%, the exciting transformer tapping was varied from half to full tap. The other variables were forward and backward phase shifts, and load and no-load conditions. Consequently, 38,400 cases of differential currents for internal faults were simulated with 960 cases for each of the 40 different types of faults.

\section{Feature Analysis}
The three-phase (a, b, and c) differential currents can be considered as time series signals. The similarity (or dissimilarity) between time series signals, e.g., a and b, can be computed at data-level using simple (e.g., Euclidean) or sophisticated (e.g., Dynamic Time Warping (DTW) \cite{DTW}) distance measures. The time series signals can also be compared at feature-level which involves computation of a set of features (e.g., mean, std, frequency, entropy, or wavelet coefficients) and the distance between the features \cite{TimeSeriesFeatureRep}. Researchers have used these features to train machine learning algorithms for classifying the signals into different categories \cite{SuccessfulFeatTimeSeries}. Nanopoulos et al. \cite{NanopoulosTimeSeriesFeat} used a variety of features from time domain which included mean, standard deviation (std), skewness, and kurtosis to classify synthetic control chart patterns used in process control. Wang et al. \cite{WangTimeSeriesFeat} extracted features from trend, seasonality, periodicity, serial correlation, skewness, kurtosis, chaos, nonlinearity, and self-similarity. The approach was further extended to multivariate time series by Wirth et al. \cite{MultiVariateStructure}. Kumar et al. \cite{KumarTimeSeries, AbenaPrimo} classified genuine and illegitimate users by computing a variety of features from time, frequency, and power domains. The specific features included band power, spectral entropy, median frequency, histogram (16 bins), range, peak magnitude to root mean square ratio, std, inter-quartile range, correlation, mutual information and DTW distances between a pair of two signals that were extracted from time series signals captured by accelerometer and gyroscope sensors. 
\begin{figure}[htp!]
\centerline{\includegraphics[width=3.3 in, height= 2.9in]{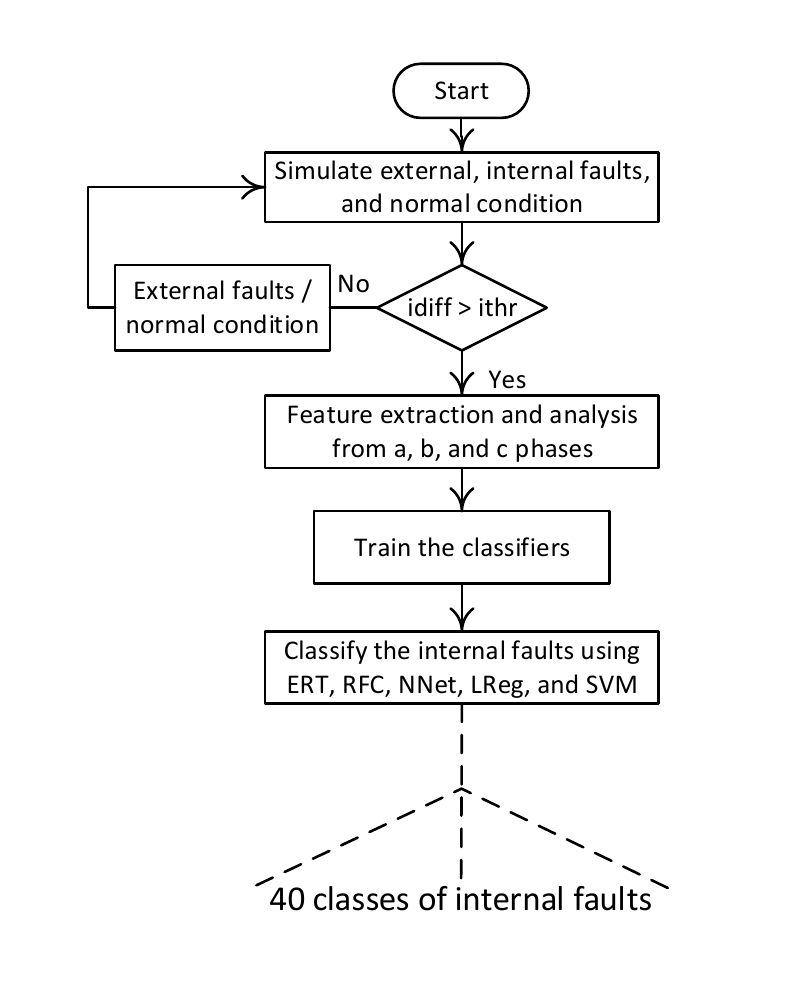}}
\vspace{-0.25in}
\caption{An illustration of the fault classification framework.}
\label{flowchart}
\end{figure}
\vspace{-0.15in}
\begin{figure*}[htp!]
\centerline{\includegraphics[width=6.8in, height= 1.6in]{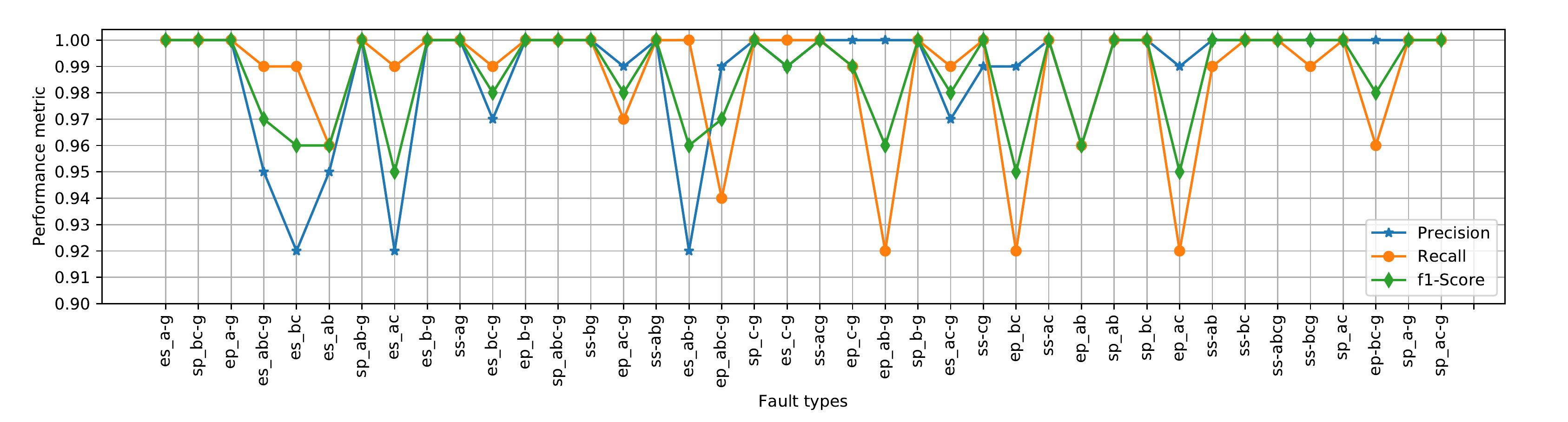}}
\vspace{-0.05in}
\caption{The performance of the best classifier i.e. ERT for each fault type was also evaluated using precision, recall, and $f_1$ scores. For a particular fault type, the precision is the ratio of the number of correctly predicted cases of that type and total number cases that are predicted of that type, i.e., $Precision = t_p/(t_p+f_p)$, where $t_p$ is true positives, $f_p$ is false positives. Similarly, the recall for a particular fault type is defined as the ratio of the number of correctly predicted cases of that type and the actual number of cases of that type, i.e., $Recall = t_p/(t_p+f_n)$, where $f_n$ is false negatives. On the other hand, $f_1$-Score is an optimal blend of both precision and recall and is defined as the harmonic mean of the two, i.e., $f_1$-Score $= 2 \times (precision \times recall)/(precision+recall)$. The higher the scores, the better the performance. The abbreviations of the fault types are as follows: ep =  exciting primary, es = exciting secondary, sp = series primary, ss = series secondary, a = a-phase, b = b-phase, c = c-phase, and g = ground.}
\label{perfclasswise}
\end{figure*}

\begin{figure*}[htp]
\centerline{\includegraphics[width=6.8in, height= 1.2in]{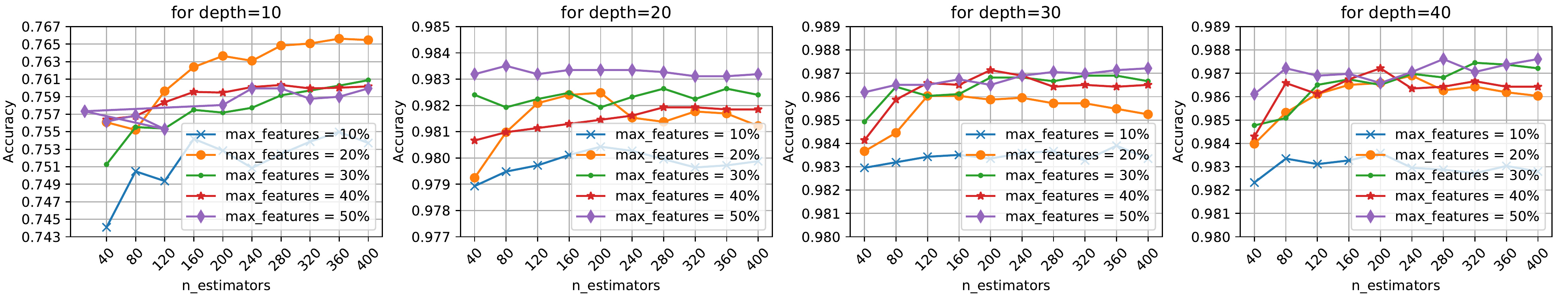}}
\vspace{-0.05in}
\caption{The performance of ERT under different settings of hyperparameter is given above. Three parameters: number of random trees (n$\_$estimators), max$\_$features per tree, and the depth of the tree, were tuned. The n$\_$estimators were varied from 40 to 400 trees with an interval of 40 trees, max$\_$features were varied from 10\% to 50\% of the total features with an interval of 10\%. More features than 50\% showed either the same or insignificant increase in the overall performances. The depth of the tree was varied from 10 to 40 with an interval of 10. The corresponding performances are presented in the Figure starting from left to right. The setting n$\_$estimators = 280, max$\_$features = 50\%, and depth=40 achieved the highest accuracy.}
\label{ertperformance}
\end{figure*}
Morchen \cite{MorchenTimeSeriesFeat} transformed the time series to wavelet and frequency domains to classify among 17 time-series datasets. Timmer et al. \cite{TimmerTimeSeriesFeat} computed features from both time and frequency domains to categorize hand tremor time series. Likewise, Bagnall et al. \cite{BagnallTimeSeriesFeat} used principal components analysis over the features extracted from the power spectrum and auto-correlation. Along the same lines, we extracted features from time domain features (e.g., min, max, mean, number of mean crossings, number of peaks, skewness, quantiles, and energy), information theoretic features (sample entropy, approximate entropy, and binned entropy), and coefficients of auto-regression, continuous wavelets, and fast Fourier transforms \cite{SuccessfulFeatTimeSeries,TSFresh}. The computation of traditional statistical features such as min, max, mean, median, quantiles, skewness, energy, and the number of peaks was straightforward and computationally inexpensive. A brief description of the rest of the features is provided in the following paragraph.

The traditional statistical features fail to capture the regularity and the unpredictability of fluctuations of time series. Therefore, the extraction of Approximate Entropy (ApEn) which measures the unpredictability of time series and has been successfully applied to identify schizophrenia, epilepsy, and addiction from EEG data. The regular the time series, the smaller the approximate entropy. ApEn, however, depends upon the length of the data and lacks relative consistency \cite{ApEnSampEn}. Consequently, the extraction of Sample Entropy (SampEn) which addresses both of these concerns but does not include self-similar patterns \cite{ApEnSampEn}. In addition to the ApEn and SampEn, the binned entropy was also extracted using a histogram of the time series. Further, the coefficients of the auto-regressive model, fast Fourier transform, and continuous wavelet transform was computed considering that the signals generated from the faults differ from each other at both time and frequency-level. These coefficients describe the signal in much more detail than the time domain features. Thus, their inclusion in our feature set. A total of 764 features were extracted from each of the three signals (a, b, and c). The influence of every feature on the fault type was evaluated by a univariate test that outputted a p-value. The lower the p-value, the more useful the feature. About 26\% of the features were removed through statistical test, resulting in an average of 564 features for each signal. 
\section{Classification Framework} Figure \ref{flowchart} illustrates the flow of fault classification framework. Internal faults were separated from external faults and normal condition by comparing the differential current (idiff) and threshold current (ithr). The features were then extracted from the data belonging to the internal faults followed by the training of identification model using five state-of-the-art classification algorithms, namely, Extremely Random Trees (ERT), Random Forest (RFC), Multilayer Perceptron (MLP), Logistic Regression (LReg), and Support Vector Machines (SVM). ERT and RFC, the ensemble-based classifiers have been shown to be effective in classifying time series signals with good accuracy \cite{KumarTimeSeries,KumarISBA}. Both of these classifiers are based on the idea of combining decision from random decision trees with some procedural difference. In the case of RFC, the random decision trees are built by using bootstrap sampling i.e. sampling with replacement from the training set. Instead of using all the features in the decision trees, a random subset of features is selected to ensure further randomization of the trees which reduces the correlation between the decisions of individual trees. The averaging of decisions from less correlated trees helps decrease the variance which improves the overall performance \cite{RFCERT}. The splitting thresholds are randomized in case of ERT which helps to reduce the variance further. There are several advantages of using the RFC and ERT classifiers including less pre-processing, quick training and testing, and relatively high accuracy compared to MLP, LReg, and SVM. The training of the identification model was carried out on 66.66\%  (640 cases for each type of faults) of the data and testing on the rest (320 cases). Python scripts were written for data prepossessing and feature analysis, whereas, the identification model was trained and tested using the scikit-learn library for Python \cite{scikit-learn}.
\begin{table}[htp]
\caption{Misclassification table illustrating the number of incorrectly predicted instances among different classes. This table does not include the fault types that had perfect identification accuracy.}
\centering
\begin{tabular}{|c|c|c|c|}
\hline
\scriptsize
\textbf{\begin{tabular}[c]{@{}c@{}}Serial\\ \#\end{tabular}} & \textbf{\begin{tabular}[c]{@{}c@{}}Actual fault\\ types\end{tabular}} & \textbf{\begin{tabular}[c]{@{}c@{}}Predicted fault\\ types\end{tabular}} & \textbf{\begin{tabular}[c]{@{}c@{}}\# of misclassified \\ cases of 320 cases \\ for each fault type\end{tabular}} \\ \hline
1                                                            & ep\_a-g                                                               & es\_a-g                                                                  & 1                                                                                                                            \\ \hline
2                                                            & es\_abc-g                                                             & ep\_abc-g                                                                & 19                                                                                                                           \\ \hline
3                                                            & es\_bc                                                                & es\_bc-g                                                                 & 2                                                                                                                            \\ \hline
4                                                            & es\_bc                                                                & ss\_ab                                                                   & 2                                                                                                                            \\ \hline
5                                                            & es\_bc                                                                & ep\_bc                                                                   & 25                                                                                                                           \\ \hline
6                                                            & es\_ab                                                                & ss\_ac-g                                                                 & 2                                                                                                                            \\ \hline
7                                                            & es\_ab                                                                & ep\_ab                                                                   & 20                                                                                                                           \\ \hline
8                                                            & es\_ac                                                                & ss\_bc-g                                                                 & 2                                                                                                                            \\ \hline
9                                                            & es\_ac                                                                & ep\_ac                                                                   & 17                                                                                                                           \\ \hline
10                                                           & ss\_a-g                                                               & ss\_bc                                                                   & 3                                                                                                                            \\ \hline
11                                                           & es\_bc-g                                                              & ep-bc-g                                                                  & 7                                                                                                                            \\ \hline
12                                                           & ep\_ac-g                                                              & es\_ac-g                                                                 & 1                                                                                                                            \\ \hline
13                                                           & es\_ab-g                                                              & ep\_ab-g                                                                 & 12                                                                                                                           \\ \hline
14                                                           & ep\_abc-g                                                             & es\_abc-g                                                                & 1                                                                                                                            \\ \hline
15                                                           & es\_c-g                                                               & ep\_c-g                                                                  & 4                                                                                                                            \\ \hline
16                                                           & ep\_ab-g                                                              & es\_ab-g                                                                 & 7                                                                                                                            \\ \hline
17                                                           & es\_ac-g                                                              & ep\_ac-g                                                                 & 7                                                                                                                            \\ \hline
18                                                           & ep\_bc                                                                & es\_bc                                                                   & 7                                                                                                                            \\ \hline
19                                                           & ep\_ac                                                                & es\_ac                                                                   & 6                                                                                                                            \\ \hline
\end{tabular}
\label{misTable}
\end{table}
\begin{table}[htp!]
\centering
\caption{The accuracy achieved by all five classifiers under different hyper parameter settings. Hyper parameters for all five classifiers were tuned using grid search. The best results are reported in this table.}
\begin{tabular}{|c|c|c|}
\hline
\textbf{Classifiers} & \textbf{Hyper parameters}                                                                      & \textbf{Accuracy} \\ \hline
\textbf{ERT}         & \begin{tabular}[c]{@{}c@{}}n$\_$estimators = 280, \\max$\_$features = 0.50,  depth=40 \end{tabular} & 98.76\%           \\ \hline
\textbf{RFC}         & \begin{tabular}[c]{@{}c@{}}n$\_$estimators=200, max$\_$depth=30,\\ max$\_$features=0.10\end{tabular} & 97.54\%           \\ \hline
\textbf{MLP}        & 2 hidden layers (300,150)                                                                      & 96.13\%           \\ \hline
\textbf{LReg}        & penalty='l2', solver='lbfgs'                                                                   & 93.54\%           \\ \hline
\textbf{SVC}         & kernel='rbf', degree=3, $\gamma$=0.001                                                            & 92.60\%           \\ \hline
\end{tabular}
\label{PerfTable}
\end{table}
\section{Results}
The identification performance was measured using accuracy, a ratio of correctly predicted instances and the total predicted instance. There was no class imbalance in the dataset as the number of cases for each type of faults was equal. The accuracy obtained by RFC, ERT, LReg, MLP, and SVM classifiers are presented in Table \ref{PerfTable}. Ensemble learning methods, ERT and RFC outperformed the rest of the classifiers achieving the identification accuracy of 98.76\% and 97.54\% respectively. The overall accuracy obtained by MLP was 96.13\%. \textit{Considering the fact that this paper analyses a total 40 different types of faults, the obtained results are significantly better than the accuracy (95.52\%) reported by Bhasker et. al \cite{pallav} over a dataset that consisted of only 20 distinct faults}.

The hyper-parameters of these classifiers were optimized using a grid search. The performance of ERT for different settings of hyperparameters are given in Figure \ref{ertperformance}. Figure \ref{perfclasswise} presents the precision, recall, and f1-score obtained by ERT for each class. The misclassification of different internal fault types is shown in Table \ref{misTable}. The rows except 3, 4, 6, 8, and 10 in the table indicate that the simulated differential currents for faults in the primary side of the exciting transformer (ep) were misclassified with the corresponding faults in the secondary side of exciting transformer (es) which is understandable as these are the two sides of the same transformer. Hence, keeping other parameters constant when the percentage of winding faulted was varied from 30\% to 70\% (in steps of 10\%) in the exciting primary (ep) and secondary (es) sides, the simulated differential fault currents confused with each other for the same kind of fault in ep and es e.g. ep$\_$bc with es$\_$bc. 

\section{Conclusion}
Two-winding and three-winding transformer components were designed to model an ISPST in PSCAD/EMTDC. Data for 40 distinct faults were generated by varying phase shift (forward and backward), fault inception angle, transformer tapping, load and no load conditions, and percentage of winding faulted. A series of features from time, frequency, time-frequency, and information theory domains were extracted and analyzed using univariate statistical tests. Five fault identifiers were implemented by employing different machine learning algorithms. The identifiers based on ensemble-based algorithms outperformed Multilayer Perceptrons, Logistic Regression, and Support Vector Machines. In the future, the analysis of different operating conditions of ISPST including over-excitation, magnetizing inrush/sympathetic inrush, turn-to-turn faults, and inter-winding faults would be considered.
\bibliography{references} 
\small{
\bibliographystyle{ieeetr}
}
\end{document}